\documentclass[letterpaper]{article} % DO NOT CHANGE THIS
\usepackage{aaai2026}  % DO NOT CHANGE THIS
\usepackage{times}  % DO NOT CHANGE THIS
\usepackage{helvet}  % DO NOT CHANGE THIS
\usepackage{courier}  % DO NOT CHANGE THIS
\usepackage[hyphens]{url}  % DO NOT CHANGE THIS
\usepackage{graphicx} % DO NOT CHANGE THIS
\usepackage{float}
\usepackage{natbib}  % DO NOT CHANGE THIS AND DO NOT ADD ANY OPTIONS TO IT
\usepackage{caption} % DO NOT CHANGE THIS AND DO NOT ADD ANY OPTIONS TO IT
\usepackage{tabularx}
\usepackage{booktabs, tabularx, pgfkeys, pgfplots}
\usepackage{multirow}
\usepackage{amsmath}
\usepackage{amsfonts}
\usepackage{subcaption}
\usepackage{makecell}
\usepackage{algorithm}
\usepackage{algorithmic}
\pgfplotsset{compat=1.18}
\usepackage{newfloat}
\usepackage{listings}
\DeclareCaptionStyle{ruled}{labelfont=normalfont,labelsep=colon,strut=off} % DO NOT CHANGE THIS
\floatstyle{ruled}
\newfloat{listing}{tb}{lst}{}
\floatname{listing}{Listing}
\title{LoKI: Low-damage Knowledge Implanting of Large Language Models}
\author{
Runyu Wang\textsuperscript{\rm 1}, 
Peng Ping\textsuperscript{\rm 2}\thanks{Corresponding author.}, 
Zhengyu Guo\textsuperscript{\rm 3}, 
Xiaoye Zhang\textsuperscript{\rm 4}, 
Quan Shi\textsuperscript{\rm 2}, 
Liting Zhou\textsuperscript{\rm 5}, 
Tianbo Ji\textsuperscript{\rm 2}
}
\affiliations{
\textsuperscript{\rm 1}School of Information Science and Technology, Nantong University\\
\textsuperscript{\rm 2}School of Transportation and Civil Engineering, Nantong University\\
\textsuperscript{\rm 3}South China University of Technology\\
\textsuperscript{\rm 4}China Southern Power Grid Company Limited\\
\textsuperscript{\rm 5}Dublin City University\\
2430310032@stmail.ntu.edu.cn, \{pingpeng, sq, jitianbo\}@ntu.edu.cn,
202264700027@mail.scut.edu.cn, xiaoyz@whu.edu.cn, liting.zhou@dcu.ie
}

\begin{document}
\maketitle

\begin{abstract}
Fine-tuning adapts pretrained models for specific tasks but poses the risk of catastrophic forgetting (CF), where critical knowledge from pretraining is overwritten. To address the issue of CF in a general-purpose framework, we propose \textbf{Lo}w-damage \textbf{K}nowledge \textbf{I}mplanting (\textbf{LoKI}), a parameter-efficient fine-tuning (PEFT) technique that utilizes recent mechanistic understanding of how knowledge is stored in transformer architectures. 
  We compare LoKI against state-of-the-art PEFT methods in two real-world fine-tuning scenarios. The results show that LoKI demonstrates significantly better preservation of general capabilities. At the same time, its task-specific performance is comparable to or even surpasses that of full parameter fine-tuning and these PEFT methods across various model architectures.
  Our work bridges the mechanistic insights of LLMs' knowledge storage with practical fine-tuning objectives, enabling an effective balance between task-specific adaptation and the retention of general-purpose capabilities.
\end{abstract}

\begin{links}
    \link{Code}{https://github.com/Nexround/LoKI}
    \link{Extended version}{https://arxiv.org/abs/2505.22120}
\end{links}

\section{Introduction}
Transformer-based language models~\cite{vaswani_attention_2017,petroni_language_2019} accumulate extensive world knowledge during pretraining~\cite{Radford2018ImprovingLU, brown_language_2020}, which becomes implicitly embedded in their parameters. Owing to their broad generalization capabilities, they serve as a strong foundation for downstream task adaptation, prompting significant efforts to develop increasingly efficient fine-tuning methods~\cite{prottasha_peft_2025,xin_parameter-efficient_2025}. 
However, the process of fine-tuning pretrained models for downstream tasks is often accompanied by catastrophic forgetting (CF)~\cite{DBLP:conf/iclr/KothaSR24,luo_empirical_2025}, a phenomenon where the model loses previously acquired capabilities after fine-tuning. Recent studies have pointed out that conventional fine-tuning approaches~\cite{prottasha_peft_2025,xin_parameter-efficient_2025} typically perform indiscriminate updates across modules within transformer architectures, oblivious to these crucial knowledge-storing weights in general tasks. 
Such unconstrained optimization may perturb crucial memory traces~\cite{cohen_crawling_2023, petroni_language_2019}, leading to irreversible knowledge loss and degraded generalization performance~\cite{li_revisiting_2024, DBLP:conf/acl/HuangCWYLSYS24}.
Extensive research has investigated the internal structure of large language models (LLMs)~\cite{DBLP:conf/aaai/Li0SYMY24, DBLP:conf/iclr/MengSABB23}, revealing that model-internal knowledge is both identifiable and editable via targeted interventions such as knowledge localization and rewriting~\cite{geva_transformer_2021-1, DBLP:conf/iclr/MengSABB23}. Moreover, the inherent redundancy in LLMs has been shown to support techniques like sparsification and parameter pruning~\cite{DBLP:conf/iclr/HeZMBN22, DBLP:conf/iclr/FrankleC19}, enabling more efficient representation without significantly degrading performance~\cite{DBLP:conf/iclr/LasbyGENI24}. However, these findings have yet to be effectively integrated into the field of model fine-tuning.
\begin{figure}[t]
\centering
\includegraphics[width=\linewidth]{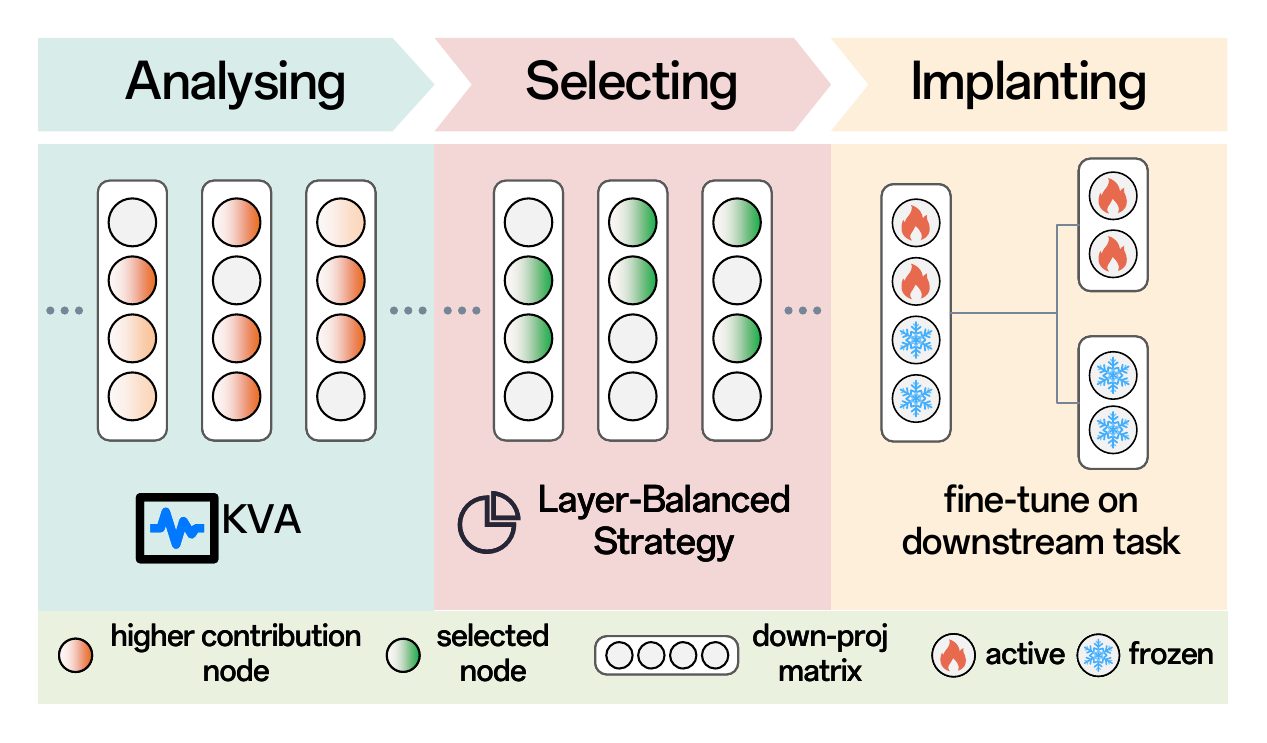}
\caption{Schematic illustration of the staged fine-tuning process in LoKI.}\label{fig:loki}

\end{figure}
These findings lead us to hypothesize that it is possible to identify low-contributing weights in pretrained models, which can then serve as capacity for injecting new, task-specific knowledge with minimal impact on existing competencies.
Based on this assumption, we propose \textbf{Lo}w-damage \textbf{K}nowledge \textbf{I}mplanting, as \textbf{LoKI}, a parameter-efficient fine-tuning method that leverages insights from interpretability studies on the internal knowledge storage mechanisms of LLMs. Aiming to mitigate catastrophic forgetting, 
LoKI consists of three stages: \textbf{analyzing}, \textbf{selecting}, and \textbf{implanting} (see Figure \ref{fig:loki}). In the \textbf{analyzing} stage, we introduce Knowledge Vector Attribution (KVA), a gradient-based attribution technique~\cite{DBLP:conf/icml/SundararajanTY17} that evaluates the contribution of each vector in the down-projection matrix $W_{\mathrm{down}}$ (see Background section for details) to the model’s pretrained behavior. In the \textbf{selecting} stage, motivated by the known interdependence between transformer layers~\cite{DBLP:conf/aaai/SunPNJ25}—especially in the progressive refinement of knowledge~\cite{geva_dissecting_2023,DBLP:conf/acl/DaiDHSCW22}—we propose the Layer-Balanced Strategy, which ensures that new knowledge aligns with the model’s hierarchical structure. Leveraging KVA results, this strategy enforces an equal quota of trainable parameters per layer by decomposing each $W_{\mathrm{down}}$ into two subsets: $W_{\mathcal{S}}$ (active) and $W_{\setminus \mathcal{S}}$ (frozen). This yields a layer-balanced trainable subset $\mathbb{W}_{\mathcal{S}}$. In the \textbf{implanting} stage, we freeze all model parameters except $\mathbb{W}_{\mathcal{S}}$, which is updated via fine-tuning.
Compared to existing fine-tuning approaches, LoKI offers several major advantages.
\begin{itemize}
    \item \textbf{Superior balance between CF and task-specific performance.} Fine-tuning LLMs of two representative sizes with LoKI effectively mitigates CF, outperforming state-of-the-art PEFT methods while maintaining strong task performance.
    \item \textbf{Intrinsically parameter-efficient.} LoKI updates only a subset of the model’s original parameters and allows for explicit control over the number of trainable weights.
    \item  \textbf{Synergistic with other tuning methods.} In addition to adapting directly for downstream tasks, we demonstrate that LoKI can be combined with existing parameter-efficient tuning techniques such as Low-Rank Adaptation (LoRA)~\cite{hu2022lora}, further reducing the number of trainable parameters.
\end{itemize}
By allocating updates to carefully selected weights, LoKI provides a competitive approach to sustainable LLM customization, enabling models to evolve while preserving their core capabilities. In summary, our main contributions are:
\begin{itemize}
\item We introduce LoKI, a fine-tuning framework for LLMs that primarily aims to mitigate CF during adaptation to downstream tasks. 
\item We propose KVA, a technique to quantify the contribution of individual knowledge vectors (defined in the Background section) to the model's stored representations. 
Furthermore, using this method, we uncover a surprising phenomenon: in transformer models, knowledge vectors with both globally high and low contributions tend to be densely located in similar layers.
\item Building on an understanding of the hierarchical organization of knowledge in transformers, we develop the Layer-Balanced Strategy for allocating trainable weights. Our experiments show that this strategy is critical to LoKI's ability to preserve pre-trained capabilities while learning new tasks.
\end{itemize}
\section{Related Works}
\label{sec:Related}
\paragraph{Catastrophic Forgetting} 
CF has long been recognized as a critical challenge in neural networks~\cite{DBLP:conf/nips/French93, kemker2018measuring}, and recent studies have begun to analyze its manifestation specifically in LLMs~\cite{li_revisiting_2024, DBLP:conf/iclr/KothaSR24}. Representative methods include: Replay-based methods~\cite{DBLP:conf/nips/dAutumeRKY19, DBLP:conf/nips/Lopez-PazR17,DBLP:conf/iclr/ChaudhryRRE19} interleave data from previous tasks during training to reinforce prior knowledge. Regularization-based methods~\cite{DBLP:journals/pami/LiH18a,DBLP:journals/corr/KirkpatrickPRVD16, deng_flattening_2021} constrain parameter updates to prevent drift from previously important weights. 
However, most of the previous methods generally treat knowledge retention as a black box, rarely considering how pretrained LLMs organize and store knowledge internally.
\paragraph{Parameter‑Efficient Fine‑Tuning} 
Recent advances in PEFT aim to mitigate CF by preserving pretrained knowledge while enabling task adaptation. For example,~\citeauthor{DBLP:conf/icml/Zhu000YD0K24} refines a small subset of critical parameters to retain performance on original tasks. However, it requires task-specific parameter selection.
Some methods structurally decouple knowledge to resist CF. CorDA~\cite{DBLP:conf/nips/YangLZSWNG24} freezes dominant singular directions, assumed to encode general knowledge, while adapting residual subspaces. LoRASculpt~\cite{DBLP:conf/cvpr/0001HWYY25} uses magnitude-based masking and conflict-aware regularization to constrain LoRA~\cite{hu2022lora} updates within critical knowledge regions.
Orthogonal subspace approaches like O-LoRA~\cite{DBLP:conf/emnlp/WangCGXBZZGH23} and LoRI~\cite{DBLP:journals/corr/abs-2504-07448} reduce task interference by isolating updates in mutually orthogonal directions, effective in multi-task scenarios but less focused on preserving original capabilities.
In contrast, LoKI proposes a one-time knowledge localization strategy, directly quantifying parameter contributions to general knowledge, enabling scalable and effective CF resistance across diverse downstream tasks.
\paragraph{Knowledge Locating and Editing} 
ROME~\cite{meng_locating_2022} revises individual factual associations by directly modifying the FFN weight vectors. KN~\cite{DBLP:conf/acl/DaiDHSCW22} regulates the expression of specific facts by controlling the activation levels of identified neurons. AlphaEdit~\cite{DBLP:conf/iclr/FangJWMSW0C25} further advances this line by applying zero-space projection to the output weight matrix of FFN layers to perform targeted edits. However, despite their effectiveness in editing specific knowledge expressions, these methods have yet to be seamlessly integrated into parameter-efficient fine-tuning pipelines that aim to retain the model’s broad pretrained capabilities. Moreover, compared to existing knowledge localization methods, KVA and the Layer-Balanced Strategy constitute a complete model analysis pipeline in practice. This combination transcends the limitation of focusing solely on specific factual expressions, evolving the underlying ideas into a general-purpose approach for model analysis.
\section{Background}
We will begin this section by briefly reviewing transformer architecture and relevant research on knowledge storage in transformer-based language models, which will establish a foundation for our proposed method. 
Transformer-based language models are built from stacked layers, each consisting of two primary components: a multi-head self-attention (MHSA) mechanism~\cite{vaswani_attention_2017} and a position-wise feed-forward network (FFN). While some studies have explored the role of MHSA in transformers~\cite{voita_analyzing_2019, DBLP:conf/blackboxnlp/ClarkKLM19,DBLP:conf/blackboxnlp/VigB19}, a growing number of studies focus on the FFN layers for knowledge localization and editing~\cite{geva_transformer_2021-1, geva_transformer_2022-1, katz_backward_2024}.
For an input vector $x \in \mathbb{R}^{d_{\mathrm{model}}}$, a typical FFN without bias applies two linear transformations with a non-linearity in between:
\begin{equation}
\mathrm{FFN}(x)
\;= W_{\mathrm{down}}\;\sigma\bigl(W_{\mathrm{up}}\,x\bigr)
\end{equation}
where $W_{\mathrm{up}} \in \mathbb{R}^{d_{\mathrm{ff}} \times d_{\mathrm{model}}}$ is the up-projection matrix, $W_{\mathrm{down}} \in \mathbb{R}^{d_{\mathrm{model}} \times d_{\mathrm{ff}}}$ is the down-projection matrix, and $\sigma(\cdot)$ denotes an element-wise activation function.
Recent studies suggest that FFN layers function as linear associative memories, where the $W_{\mathrm{up}}$ acts as a collection of keys detecting input patterns, and the $W_{\mathrm{down}}$ contains values corresponding to interpretable concepts~\cite{geva_transformer_2021-1, geva_transformer_2022-1}. This view has been reinforced by both activation- and gradient-based analyses~\cite{katz_backward_2024}. Specifically, each output coordinate $y_j$ in the FFN can be viewed as a knowledge output node, computed as $y_j = v_j^\top \sigma(W_{\mathrm{up}},x)$, where $v_j$ is the $j$-th row of $W_{\mathrm{down}}$ and serves as a knowledge vector.  
On the other hand, while single-layer FFN reveals this memory behavior, deeper inspection shows a hierarchy: lower FFN layers capture surface-level patterns, whereas upper layers encode higher-level semantics~\cite{geva_dissecting_2023,DBLP:conf/acl/DaiDHSCW22,DBLP:conf/iclr/TanZF24,katz_backward_2024}. The final output distribution of transformer-based language models is gradually constructed in a bottom-up fashion~\cite{DBLP:conf/aaai/SunPNJ25,tenney_bert_2019,wallat_bertnesia_2020}.
Such insights have inspired a wave of model editing techniques that target the FFN layers—particularly $W_{\mathrm{down}}$—to insert, update, or erase factual knowledge without extensive retraining~\cite{geva_transformer_2022-1, dai_knowledge_2022-1,meng_mass-editing_2023}. Motivated by the success of these approaches, our study focuses on implanting task-specific knowledge through FFNs, which we consider the main site for implanting new knowledge into the model.
Importantly, prior research reveals that LLMs exhibit considerable parameter redundancy, especially within FFNs~\cite{DBLP:conf/iclr/KobayashiKYI24, DBLP:conf/nips/Sanh0R20,DBLP:conf/acl/ZhangL00S022,sanh_movement_2020,DBLP:conf/icml/FrantarA23}. Numerous studies show that substantial portions of model weights can be pruned or restructured with minimal performance loss on general tasks~\cite{DBLP:conf/iclr/FrankleC19, DBLP:conf/nips/MichelLN19, sanh_movement_2020, DBLP:conf/icml/FrantarA23}. 
Based on the previous research mentioned above, we hypothesize that these low-impact parameters in FFN layers can be repurposed to encode new task-specific knowledge without excessively degrading the model's original capabilities. Our experimental results provide evidence supporting this hypothesis.
\section{Low-damage Knowledge Implanting}
\label{headings}
We introduce LoKI, a three-stage framework—\textbf{analyzing}, \textbf{selecting}, and \textbf{implanting}—designed to edit transformer models with minimal disruption, thereby mitigating CF (see Figure \ref{fig:loki}). Below, we briefly outline each stage before providing a detailed explanation:
\begin{enumerate}
\item \textbf{Analyzing}: Evaluate the contribution of each knowledge vector to general tasks.
\item \textbf{Selecting}: Choose trainable knowledge vectors within each FFN based on the analysis results.
\item \textbf{Implanting}: Train the chosen vectors to incorporate task-specific knowledge.
\end{enumerate}
\subsection{Analysing}
\label{sec:Analysing}
We start this subsection by introducing \textbf{Knowledge Vector Attribution (KVA)}, an attribution method inspired by~\cite{DBLP:conf/aaai/Hao0W021,DBLP:conf/acl/DaiDHSCW22}, designed to evaluate the contribution of individual knowledge vectors to the model's performance. Next, we illustrate the workflow of the analysis stage—in other words, how KVA is utilized within the LoKI framework.
\subsubsection{Knowledge Vector Attribution}
KVA is a computational approach based on Integrated Gradients (IG)~\cite{DBLP:conf/icml/SundararajanTY17} that measures the contribution of knowledge vectors to specific output logits. 
\begin{equation}
\mathbf{IG}_i(x) = (x_i - x'_i) \int_0^1 \frac{\partial F(x'+\alpha(x-x'))}{\partial x_i} d\alpha,
\end{equation}
where $F$ is the network function, $x'$ is a baseline (we use \(x' = \mathbf{0}\)), and $\alpha \in [0,1]$ is an interpolation coefficient that defines the integration path from baseline to input. IG attributes predictions to input features by integrating gradients along the straight-line path from baseline $x'$ to input $x$.
To trace knowledge flow through all layers of a transformer on input sequence \(\mathbf{x}\), let \(\mathcal{L}\) denote the target logit, \(h_{l-1}\) represent the hidden state from the previous layer. At layer \(l\), the FFN first computes pre-activations
\begin{equation}
z_{l} = W_{\mathrm{up}}^{(l)}\,h_{l-1},
\quad
\mathbf{u}_{l} = \sigma(z_{l}),
\quad
\mathbf{z}_{l,j} = \mathbf{u}_{l,j}\,W_{\mathrm{down},j\cdot}^{(l)}.
\end{equation}
As mentioned above, we treat FFNs as collections of key-value memory pairs. To attribute the contribution of each knowledge vector to the final output of an LLM, we define the layer-wise, path-integrated attribution of node \(j\) by
\begin{equation}
Attr_{l,j}(\mathbf{x})
= \int_{0}^{1}
    \frac{\partial \mathcal{L}\bigl(\alpha\,\mathbf{z}_{l,j}\bigr)}
         {\partial \mathbf{z}_{l,j}}
  \,d\alpha,
\end{equation}
which aggregates the total gradient flowing through knowledge vector \(j\) as its contribution is scaled from zero up to its actual activation. We approximate the above integral via Riemann approximation with \(m\) equally-spaced steps (we set the step $m=7$ for any model, and further discussion on the setting of \(m\) can be found in Appendix A):
\begin{equation}
Attr_{l,j}(\mathbf{x})
\approx \frac{1}{m}
\sum_{k=1}^{m}
  \frac{\partial \mathcal{L}\bigl(\tfrac{k}{m}\,\mathbf{z}_{l,j}\bigr)}
       {\partial \mathbf{z}_{l,j}}.
\end{equation}
During each sample, KVA computes and stores $Attr_{l,j}(\mathbf{x})$ for every layer $l$ and every knowledge output node $j$. This yields a complete attribution log that can be analyzed post hoc to assess the contribution of each knowledge vector during inference.
\subsubsection{Scoring Knowledge Vectors}
Pretrained LLMs exhibit a wide range of capabilities, including world knowledge, common-sense reasoning, and instruction following. While it is difficult to precisely determine the full scope of an LLM’s internal knowledge, it can be approximated through extensive textual input~\cite{DBLP:conf/iclr/MengSABB23}.
To evaluate the contribution of individual knowledge vectors to general-purpose performance—and to identify those with minimal impact on core capabilities—a critical factor is the choice of dataset for quantitative analysis.
To ensure broad domain coverage, we apply KVA to the \texttt{Massive Multitask Language Understanding (MMLU)} benchmark~\cite{DBLP:conf/aaai/KemkerMAHK18}. MMLU is ideal because:
\begin{itemize}
  \item it has 57 diverse subjects (STEM, humanities, professional, etc.), which covers a broad spectrum of general-knowledge tasks, ensuring the evaluation score of vectors is not domain-specialized;
  \item its standardized prompts and evaluation make attributions comparable across tasks;
  \item high performance on MMLU correlates with real-world language understanding, so protecting original performance is measurable.
\end{itemize}
This process, executed on an RTX4090 GPU using Llama3.1-8B-Instruct~\cite{DBLP:journals/corr/abs-2407-21783}, takes an average of 9.69 seconds per sample.
In parallel, we apply KVA to the full MMLU dataset. Results show that the final selected nodes—based on the Layer-Balanced Strategy (detailed later)—overlap by \textbf{97.57\%} between the reduced set and the full dataset. This confirms that our sampling strategy substantially reduces computational cost while preserving high fidelity in attribution outcomes.
Notably, KVA needs to be performed only once per model, independent of downstream tasks. That is, this computational overhead is incurred a single time per model. Implementation details and additional discussion are provided in Appendix A.
\subsection{Selecting}
\label{sec:selecting}
\begin{figure}[!h]
    \centering
    \includegraphics[width=\linewidth, trim=0 10 0 0, clip]{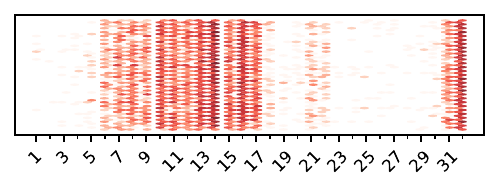}
    \includegraphics[width=\linewidth, trim=0 10 0 0, clip]{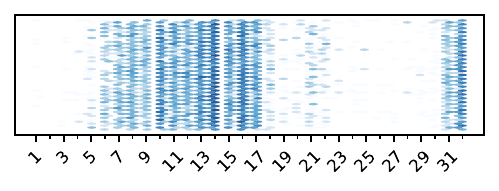}
    \caption{Heatmaps of the top $5\%$ KVA results across all 32 layers of Llama3.1-8B-Instruct. The vertical axis denotes node indices, and the horizontal axis denotes layer indices. The upper (red-tinted) heatmap illustrates the distribution of high-contribution node positions, while the lower (blue-tinted) heatmap illustrates the distribution of low-contribution node positions. Color intensity (log-scale) reflects the density of nodes within each category, with darker colors indicating higher density. Heatmaps for additional models are provided in Appendix B.}
    \label{fig:kva}
\end{figure}
As shown in Figure \ref{fig:kva}, the evaluation results of KVA reveal significant heterogeneity in the distribution of knowledge output nodes across different layers, regardless of their contribution level. This uneven distribution phenomenon among nodes with high contributions aligns with the findings of Meng et al. \shortcite{meng_locating_2022, DBLP:conf/iclr/MengSABB23} and Dai et al. \shortcite{dai_knowledge_2022-1}. 
Notably, our results uncover a novel phenomenon: both high- and low-contribution nodes are densely concentrated in the same layers. This suggests a potential structural relationship in the placement of knowledge vectors across different attribution levels. We believe this observation may offer new insights into inter-layer knowledge organization in transformer architectures. As our current focus is on leveraging the analysis results for guiding fine-tuning, we leave a deeper investigation of this phenomenon to future work.
These findings shed a critical light on fine-tuning strategies: the layer-wise distribution of knowledge is not random but structurally biased. Consequently, naïve parameter-efficient approaches that update only a limited number of layers may inadvertently disrupt the intrinsic knowledge hierarchy of transformers~\cite{sun_transformer_2025,geva_transformer_2022-1}. Our experiments show that such imbalanced parameter allocation exacerbates catastrophic forgetting (see the Ablation Studies section). Furthermore, Hase et al. \shortcite{DBLP:conf/nips/HaseBKG23} have found a significant challenge: even when factual knowledge resides in specific layers, such as the mid-layer FFNs, modifying weights in distant layers, especially the earlier ones, can effectively "override" the information flow downstream.
\subsubsection{Layer-Balanced Strategy} Based on the previous considerations, we propose the Layer-Balanced Strategy, which is a strategy that determines the trainable parameter positions under the constraint of allocating the same number of parameters to each layer. This strategy aims to (a) ensure that newly implanted knowledge conforms to the hierarchical relationship of the transformer model's knowledge structure, and (b) avoid disproportionately disturbing its inherent knowledge hierarchy.
Let the model have $L$ layers, where only parameters within each layer's $W_{\mathrm{down}}$ matrix are partially trainable, while all other parameters across all layers remain frozen. Each $W_{\mathrm{down}}$ in layer $l$ contains $D_l$ knowledge output nodes (with $D_l = D$ assumed identical across layers for simplicity). The hyperparameter $q \in (0,100)$ governs the percentage of trainable nodes selected from all $W_{\mathrm{down}}$ matrices. Given $N$ inference samples, we denote the KVA result for node $i$ in layer $l$ on sample $t$ as:
$$
\mathbf{Attr}_{l,i}^{(t)}, \quad i=1,\dots,D,\; t=1,\dots,N
$$
The implementation proceeds as follows:
\begin{enumerate}
    \item \textbf{Quota Allocation:} 
    Calculate the total trainable slots:
    \begin{equation}
      T = \frac{q}{100} \cdot \sum_{l=1}^L D_l = \frac{q}{100} \cdot L D
    \end{equation}
    Allocate equal quotas per layer: 
    \(
    k_l = \left\lfloor \frac{T}{L} \right\rfloor, \quad l=1,\dots,L. 
    \)
  \item \textbf{Per-Sample Local Selection:}
    For each sample $t$ and layer $l$, select $k_l$ nodes with the smallest values:
    \begin{equation}
      S_l^{(t)} = \underset{i\in\{1,\dots,D\}}{\mathrm{argsort}\downarrow}\left(\hat{\mathbf{Attr}}_{l,i}^{(t)}\right)[1:k_l]
    \end{equation}
  \item \textbf{Frequency Aggregation:}
    Tally selection frequencies across samples:
    \begin{equation}
      c_{l,i} = \sum_{t=1}^N \mathbb{I}\left(i \in S_l^{(t)}\right)
    \end{equation}
  \item \textbf{Final Selection:}
    For each layer $l$, select nodes with the highest frequencies:
    \begin{equation}
      \mathcal{S}_l = \underset{i\in\{1,\dots,D\}}{\mathrm{argsort}\uparrow}(c_{l,i})[1:k_l], \quad \mathcal{S} = \bigcup_{l=1}^L \mathcal{S}_l
    \end{equation}
    Resulting in a balanced set $\mathcal{S}$ with $\sum_{l=1}^L |\mathcal{S}_l| \leq T$.
\end{enumerate}
To empirically assess the effectiveness of the Layer-Balanced Strategy, we perform an ablation study comparing it with imbalanced baselines that naïvely select knowledge vectors with the highest or lowest global attribution scores as trainable parameters. Experimental details and results are provided in the Ablation Studies section.
\subsection{Implanting}
\label{sec:implanting}
Building upon the selected knowledge output nodes $\mathcal{S} = \bigcup_{l=1}^L \mathcal{S}_l$, we implement parameter-efficient fine-tuning through strategic decomposition of FFN layers. For each layer $l$'s down-projection matrix $W_{\mathrm{down}}^{(l)}$, we partition the parameters into two complementary subspaces:
\begin{equation}
W_{\mathrm{down}}^{(l)} = \begin{bmatrix}
W_{\mathcal{S}_l} \\[3pt]
W_{\setminus \mathcal{S}_l}
\end{bmatrix},
\end{equation}
where $W_{\mathcal{S}_l} \in \mathbb{R}^{|\mathcal{S}_l| \times d_{\mathrm{ff}}}$ contains the weights corresponding to our selected low-contribution knowledge output nodes (from layer $l$'s quota $\mathcal{S}_l$), and $W_{\setminus \mathcal{S}_l}$ represents the remaining parameters. During training, we:
\begin{itemize}
    \item Keep $W_{\setminus \mathcal{S}_l}$ \textbf{frozen} to preserve existing knowledge representations
    \item Update only $W_{\mathcal{S}_l}$ \textbf{actively} to implant new knowledge
\end{itemize}
This decomposition retains the mathematical formulation of the original layer while restricting parameter updates to the chosen subspaces. In addition, it transforms LoKI training into a module-wise process, allowing easy integration with existing training pipelines. For clarity, we refer to this implementation as \textbf{LoKI Linear}, and the readers can find its PyTorch code in Appendix C.
\subsubsection{Incorporating LoRA}
It is worth noting that this implementation allows for the incorporation of low-rank decomposition techniques~\cite{hu2022lora,DBLP:conf/icml/LiuWY0WCC24, DBLP:conf/iclr/ZhangCBH0CZ23} into LoKI. Specifically, the trainable weights are parameterized as:
\begin{equation}
W_{\mathcal{S}_l} = W_{\mathcal{S}_l}^{(0)} + \Delta W_{\mathcal{S}_l}, \quad \Delta W_{\mathcal{S}_l} = A_l B_l,
\end{equation} where $W_{\mathcal{S}_l}^{(0)}$ denotes the frozen base weights, and $A_l \in \mathbb{R}^{|\mathcal{S}_l| \times r}$, $B_l \in \mathbb{R}^{r \times d_{\mathrm{ff}}}$ are learnable low-rank matrices with rank $r \ll \min(|\mathcal{S}_l|, d_{\mathrm{ff}})$.
To verify the feasibility of this integration, we experimented with the Experiment section.
\section{Experiments}
\label{Experiments}
To evaluate our proposed method, we focus on adapting models to the following two datasets: 
\begin{enumerate}
    \item \textbf{ToolACE Function-Calling Dataset}~\cite{liu_toolace_2024}: This dataset contains 26,507 distinct APIs across various domains, designed to enhance the function-calling capabilities of LLMs. Consistent with the official model released by the dataset provider, we fine-tuned Llama3.1-8B-Instruct on this dataset. The details of the experiment setup can be found in Appendix D.
    \item \textbf{LB Reranker Dataset}~\cite{lb_reranker}: This multilingual dataset consists of 2.28 million query-text pairs annotated with fine-grained relevance scores on a 1-7 scale, designed for training NLP-based retrieval models. It has an official full parameter fine-tuning model based on Qwen2.5-0.5B-Instruct~\cite{qwen_qwen25_2025}; we utilized the same base model on this dataset. The experiment setup for this task can be found in Appendix E.
\end{enumerate}
These publicly available datasets were specifically selected due to their demonstrated practical utility in real-world applications and their ability to simulate realistic fine-tuning demand scenarios. Each experiment assesses the capability of LoKI to resist the CF phenomenon while acquiring task-specific performance. Notably, while our experiments involved only two model types, the proposed method is readily applicable to other model architectures.

\begin{table*}[!t]\linespread{0.75}\selectfont
\centering

\small{
\begin{tabular}{l|c|cc|c|cc}
\toprule
\multirow{2}{*}{Model} 
& \multirow{2}{*}{\makecell{Overall Acc\\(\%, ↑)}} 
& \multicolumn{2}{c|}{\makecell{Single Turn Acc}}     
& \multirow{2}{*}{\makecell{Multi Turn\\(\%, ↑)}} 
& \multicolumn{2}{c}{\makecell{Hallucination Measurement}} \\
\cline{3-4} \cline{6-7}
&   & \makecell{Non-Live(\%, ↑)} & \makecell{Live(\%, ↑)} 
&                             & \makecell{Relevance(\%, ↑)} & \makecell{Irrelevance(\%, ↓)} \\
\midrule
ToolACE$\dagger$($r{=}16$)     & 58.32 & \textbf{87.56} & \textbf{76.10} & 7.62  & 83.33 & 88.05 \\
DoRA($r{=}16$)  & \underline{58.90} & 82.04 & 75.61 & 16.00  & 83.33 & 88.92 \\
PiSSA($r{=}16$)  & 53.97    & 80.56 & 72.68 & 5.62  & 61.11 & 92.19  \\
\midrule
LoKI($q{=}10$)   & 56.76             & 80.81          & 70.46          & \underline{16.50} & 83.33 & \textbf{82.21} \\
LoKI($q{=}20$)   & 58.16             & 81.02          & 73.26           & \textbf{17.75}    & 83.33 & 85.73 \\
LoKI($q{=}30$)   & \textbf{58.93}    & \underline{82.71} & \underline{75.70} & 16.25          & 83.33 & 87.65 \\
LoKI*($q{=}30,r{=}32$)   & 57.16    & 81.96 & 71.66 & 15.75  & 83.33 & \underline{84.01}  \\
\bottomrule
\end{tabular}
}
\caption{Performance comparison on the Berkeley Function Calling Leaderboard V3. ToolACE refers to the official model trained using LoRA. $r$ denotes the rank. Models marked with $\dagger$ indicate that the corresponding performance metrics are sourced directly from the official BFCL documentation. An asterisk (*) denotes models where LoRA was applied during training. \textbf{Bold} represents the top performance score in each column; \underline{underline} represents the runner-up.}\label{tab:toolace}
\vspace*{1em}
\small{
\begin{tabular}{l|c|cccccc|r}
\toprule
Model & \#Params & TriviaQA & GSM8K & Hellaswag & WinoGrande & HumanEval & IFEval & Avg(↓) \\
\midrule
Llama3.1 (untuned)        & --       & 65.77  & 84.46  & 73.85 & 62.98 & 68.29 & 79.76 & -- \\
\midrule
ToolACE($r{=}16$)   & 42M       & 64.63  & 79.53 & 46.81 & 42.78 & 60.37 & 72.76 & 16.11\% \\
DoRA($r{=}16$)      & 43M      & 63.97  & 82.64 & 70.78 & 57.85 & 65.24 & 73.47 & 4.92\% \\
PiSSA($r{=}16$)     & 42M      &  52.50 & 51.40 & 37.98 & 13.18 & 15.85 & 55.95 & 48.93\% \\
\midrule
LoKI($q{=}10$)  & 188M      & 65.85  & 84.99 & 75.14 & 61.33 & 68.90 & 77.54 & 0.34\% \\
LoKI($q{=}20$)  & 376M      & 65.60  & 84.31 & 73.60 & 60.62 & 67.68 & 78.18 & 0.93\% \\
LoKI($q{=}30$)  & 563M      & 65.49  & 84.61 & 71.16 & 61.09 & 70.73 & 77.94 & 1.23\% \\
LoKI*($q{=}30,r{=}32$) & 16M       & 64.30  & 83.47 & 70.78 & 62.51 & 70.12 & 78.21 & 1.26\% \\
\bottomrule
\end{tabular}
}

\caption{Benchmark scores of models fine-tuned on ToolACE Function-Calling Dataset. Llama3.1 denotes Llama3.1-8B-Instruct. Avg denotes the average performance degradation percentage of each indicator compared to the original model indicator. We complete all the benchmarks on OpenCompass and report their results directly.}\label{tab:toolace_benchmark}
\end{table*}

\begin{table}[h]\linespread{0.75}\selectfont
\centering
{\small
\begin{tabularx}{\linewidth}{l|*{4}{>{\centering\arraybackslash}X}}
\toprule
Model & MAP@1 (\%) & Recall@1 (\%) & NDCG@1 (\%) & P@1 (\%) \\
\midrule
Qwen2.5& -78.1 & -78.1 & -78.7 & -77.5 \\
\midrule
DoRA($r{=}8$)     & -8.7 & -8.7 & -11.2 & -10.4 \\
PiSSA($r{=}8$)    & -9.6  &  -9.6 &  -9.7  &  -10.2  \\
CorDA($r{=}8$)    & -3.3  &  -3.1  &  -2.6  & -3.3 \\
\midrule
LoKI($q{=}5$)     & -3.1  & -3.1  & -2.2  & -1.8  \\
LoKI($q{=}10$)    & -2.3  & -2.3  & -0.3  & -0.5  \\
LoKI($q{=}20$)    & \underline{+0.2} & \underline{+0.2} & \underline{+0.7} & \underline{+0.8} \\
LoKI($q{=}30$)    & \textbf{+1.0} & \textbf{+1.0} & \textbf{+2.1} & \textbf{+2.5} \\
\bottomrule
\end{tabularx}
}
\caption{Retrieval performance on BEIR benchmark. Results show the percentage difference relative to the full parameter fine-tuning model across standard retrieval metrics, where positive values indicate superior performance. CorDA uses the KPM setting on the NQ-Open dataset.}\label{tab:reranker_beir}
\end{table}

% table 2
% \begin{table*}[h!]
% \centering
% \end{table*}

\paragraph{Measuring Catastrophic Forgetting}
To systematically evaluate CF during fine-tuning, we assess the model’s retained general capabilities across six diverse benchmarks: \texttt{TriviaQA}~\cite{DBLP:conf/acl/JoshiCWZ17} (world knowledge), \texttt{GSM8K}~\cite{cobbe2021gsm8k} (mathematical reasoning), \texttt{HellaSwag}~\cite{DBLP:conf/acl/ZellersHBFC19} and \texttt{WinoGrande}~\cite{DBLP:conf/aaai/SakaguchiBBC20} (commonsense understanding), \texttt{HumanEval}~\cite{chen2021evaluating} (code generation), and \texttt{IFEval}~\cite{zhou_instruction-following_2023} (instruction following). 

We define the average forgetting score across all benchmarks as:
$
\text{Avg} = \frac{100}{N} \sum_{i=1}^{N} \frac{S_i^{\text{o}} - S_i^{\text{t}}}{S_i^{\text{o}}},
$
where $S_i^{\text{o}}$ and $S_i^{\text{t}}$ denote the performance of the original and fine-tuned models on the $i$-th benchmark, respectively, and $N$ is the total number of benchmarks. A higher value of $\text{Avg}$ indicates a greater degree of forgetting. Further details regarding the evaluation of these benchmarks can be found in Appendix~F.
\subsubsection{Task 1: ToolACE Function-Calling Dataset}
\label{sec:toolace}
\label{toolace}

\begin{table*}[!ht]\linespread{0.75}\selectfont
\centering
\small{
\begin{tabular}{l|c|cccccc|r}
\toprule
Model & \#Params & TriviaQA & GSM8K & HellaSwag & WinoGrande & HumanEval & IFEval & Avg(↓) \\
\midrule
Qwen2.5 (untuned)   & --          & 24.37 & 39.65 & 30.98 & 44.20 & 26.83 & 33.39 & -- \\
\midrule
LB-Reranker         & 494M        & 4.65  & 2.65  & 0.00  &   0.00  & 0.00  & 23.20 & 84.13\%  \\
DoRA($r{=}8$)         & 4M        & 19.30& 34.01& 22.21& 44.70& 25.22& 31.67 & 12.23\% \\
PiSSA($r{=}8$)         & 4M       & 11.32 & 6.44 & 18.54 & 31.97 & 6.71 & 28.85 &  48.95\% \\
CorDA($r{=}8$)         & 4M       & 4.64 & 2.43 & 0.00 & 0.00 & 0.00 &  23.20&  84.22\% \\
\midrule
LoKI($q{=}5$)       & 5.1M        & 20.53 & 38.36 & 30.52 & 40.09 & 24.39 & 34.12 & 6.12\% \\
LoKI($q{=}10$)      & 10.4M       & 20.53 & 37.98 & 21.53 & 47.04 & 24.39 & 33.39 & 8.86\% \\
LoKI($q{=}20$)      & 20.9M       & 19.77 & 36.77 & 33.16 & 47.28 & 27.44 & 33.27 & 1.70\% \\
LoKI($q{=}30$)      & 31.3M       & 20.10 & 37.60 & 38.39 & 43.96 & 26.83 & 32.25 & 0.46\% \\
\bottomrule
\end{tabular}
}
\caption{Benchmark scores of models fine-tuned on the LB Reranker Dataset. LB-Reranker refers to the full parameter fine-tuning model released by the dataset provider. Qwen2.5 represents the Qwen2.5-0.5B-Instruct.}\label{tab:reranker_benchmark}
\end{table*}

Based on the training results on this dataset, we first compare LoKI with existing SOTA methods in terms of downstream task adaptation. We evaluate fine-tuned models' performance on the Berkeley Function Calling Leaderboard V3, comparing LoKI against LoRA (model provided by the dataset authors), DoRA~\cite{DBLP:conf/icml/LiuWY0WCC24}, and PiSSA~\cite{DBLP:conf/nips/MengWZ24}. As shown in Table~\ref{tab:toolace}, LoKI achieves the highest overall accuracy when $q{=}30$. Under the $q{=}20$ setting, LoKI exhibits the strongest multi-turn reasoning capability, with a success rate of \textbf{17.75\%}. Notably, all LoKI variants consistently reduce the Irrelevance metric, suggesting that LoKI may help mitigate hallucination effects introduced during fine-tuning.
Table \ref{tab:toolace_benchmark} further reveals LoKI's unique capability in preserving foundational model competencies. Compared to DoRA, LoKI ($q{=}30$) not only outperforms in fine-tuned performance, but also reduces the average performance degradation percentage by \textbf{75\%} across six evaluation benchmarks. Relative to the original pretrained model, the degradation is as small as \textbf{1.23\%}. Notably, as the hyperparameter $q$ increases from 10 to 30—despite a substantial growth in the number of trainable parameters—the rate of performance degradation slows down.
Additionally, we experimented to investigate the potential of combining LoRA with LoKI (LoKI*($q{=}30$)). As anticipated, the number of trainable parameters in the model significantly decreased when integrated with LoRA, showing a reduction of 97.16\% compared to LoKI($q{=}30$). Importantly, this combination did not visibly compromise LoKI's ability to resist catastrophic forgetting. we believe that this combination holds significant promise with further optimization of training settings.

\subsubsection{Task 2: LB Reranker Dataset}

The progressive performance improvement with increasing trainable parameters (from $q{=}5$ to $q{=}30$) reveals a clear parameter-performance tradeoff. Even with only $q{=}20$, LoKI already achieves positive performance gains in four metrics. 
As shown in Table \ref{tab:reranker_benchmark}, compared to all baseline methods including DoRA, PiSSA, and CorDA~\cite{DBLP:conf/nips/YangLZSWNG24}, LoKI models of all configurations exhibit significantly less performance degradation across all metrics. Interestingly, when $q{=}30$, LoKI not only achieved the best averaged performance on the BEIR benchmark, but also showed the least degradation on general tasks. We attribute this phenomenon to our empirical learning rate schedule: higher $q$ values used proportionally lower learning rates (e.g., the learning rate for $q{=}30$ was 0.4$\times$ that for $q{=}10$), which appears to provide a better balance between task adaptation and knowledge preservation. Further exploration of this hyperparameter interplay is discussed in Appendix E.

\section{Ablation Studies}
\label{sec:ablation}

\begin{table}[!ht]\linespread{0.75}\selectfont

\centering
\begin{subtable}[c]{0.48\linewidth}
\centering
\caption{}
\label{tab:ablation_kva}
{\small

\begin{tabular}{l|r}
\toprule
Model & Avg(↓) \\
\midrule
S-H   & 36.73\% \\
S-L   & 11.35\% \\
\bottomrule
\end{tabular}
}
\end{subtable}
\hfill
\begin{subtable}[c]{0.48\linewidth}
\centering
\caption{}
\label{tab:ablation_lbs}
{\small

\begin{tabular}{l|r}
\toprule
Model & Avg(↓) \\
\midrule
LoKI & 8.86\% \\
G-H  & 39.04\% \\
G-L  & 30.48\% \\
\bottomrule
\end{tabular}
}
\end{subtable}
\caption{
(a) Performance of two suppression strategies on benchmarks when $q{=}1$. 
(b) Performance comparison of different methods when $q{=}10$. 
Detailed results are provided in Appendix G.
}
\end{table}

In this section, we evaluate (a) the effect of KVA in quantifying knowledge vectors' contributions to general task performance, and (b) the necessity of the Layer-Balanced Strategy in LoKI.
\subsubsection{Validation of KVA}
\label{sec:ablation_kva}
We use Llama3.1-8B-Instruct to validate if KVA can effectively attribute the contribution of knowledge vectors to general tasks. We compare two suppression strategies (i.e., zeroing the corresponding weights) with the base model using the same benchmarks in the Experiments section:
\begin{itemize}
\item \textbf{S-H/L}: Suppresses the top $q$\% knowledge vectors that appear most frequently with the \textbf{highest}/\textbf{lowest} attribution scores (expected to \textbf{significantly}/\textbf{minimally} degrade performance if KVA correctly identifies critical vectors).
\end{itemize}
As presented in Table \ref{tab:ablation_kva}, when the top 1\% of high-contribution vectors are suppressed, there is a significant performance decline, with an average degradation of 36.73\%. In contrast, the performance gap between the two suppression strategies is \textbf{25.38\%}, which indicates significant differences and supports the accuracy of KVA.
\subsubsection{Validation of Layer-Balanced Strategy}
\label{sec:ablation_lbs}
Next, we test the necessity of the Layer-Balanced Strategy by fine-tuning Qwen2.5-0.5B-Instruct on the LB Rerank Dataset, using two globally imbalanced schemes. Equally, we compare these results with models trained using LoKI on the 6 benchmarks mentioned above.
\begin{itemize}
  \item \textbf{G-H/L}: Globally set the top $q$\% knowledge vectors across all layers that appear most frequently with the \textbf{highest}/\textbf{lowest} attribution scores as trainable parameters.
\end{itemize}
As shown in Table~\ref{tab:ablation_lbs}, both imbalanced strategies fall short of LoKI, despite using the same quota, underscoring the importance of the Layer-Balanced Strategy. Additional BEIR benchmark results are included in Appendix~G.
\section{Conclusions}
We present LoKI, a parameter-efficient fine-tuning framework that achieves balanced adaptation between downstream task performance and preservation of pre-trained knowledge in large language models. Our key insight stems from systematically analyzing the hierarchical knowledge storage mechanism in transformers and developing a layer-balanced parameter selection strategy guided by integrated gradient attribution. Through experiments on retrieval and tool-use tasks, we demonstrate that LoKI achieves competitive task adaptation while significantly reducing catastrophic forgetting compared to full parameter fine-tuning and prevalent PEFT methods. Our experimental results demonstrate that integrating insights from mechanistic interpretability research with fine-tuning objectives is effective, highlighting the potential of this interdisciplinary direction.

\section{Acknowledgement}
This work was supported in part by the National Natural Science Foundation of China under Grants 52202496, 52442218, and U2433216; The Key Research and Development Project of Nantong City, China (Special Project for Prospective Technology Innovation, No. GZ2024001); and the Key Laboratory of Target Cognition and Application Technology (2023-CXPT-LC-005).
\bibliography{aaai2026}
\newpage

\appendix
\setcounter{secnumdepth}{2}
\renewcommand{\thesection}{\Alph{section}}
\renewcommand{\thesubsection}{\Alph{section}.\arabic{subsection}}
\renewcommand{\thesubsubsection}{\Alph{section}.\arabic{subsection}.\arabic{subsubsection}}

\section{Details and Discussions of KVA}
\label{appendix_a}
\subsection{Exploration of Different Settings}
\subsubsection{Step setting}

\begin{table}[h!]
\centering
\begin{tabular}{c|c||c|c}
\toprule
\textbf{Layer} & \textbf{Sim. (\%)} & \textbf{Layer} & \textbf{Sim. (\%)} \\
\midrule
0  & 83.15  & 12 & 98.88 \\
1  & 94.38  & 13 & 98.88 \\
2  & 97.75  & 14 & 100.00 \\
3  & 96.63  & 15 & 95.51 \\
4  & 98.88  & 16 & 98.88 \\
5  & 97.75  & 17 & 97.75 \\
6  & 97.75  & 18 & 100.00 \\
7  & 95.51  & 19 & 98.88 \\
8  & 98.88  & 20 & 100.00 \\
9  & 98.88  & 21 & 98.88 \\
10 & 97.75  & 22 & 97.75 \\
11 & 96.63  & 23 & 98.88 \\
\midrule
\multicolumn{3}{r|}{\textbf{Total}} & \textbf{97.42} \\
\bottomrule
\end{tabular}
\caption{The similarity of each layer and the total similarity of trainable nodes obtained using two different Riemann approximation steps on Qwen2.5-0.5B-Instruction. Sim. represents similarity.}
\label{tab:qwen_sim_step}
\end{table}

We examined the effect of the Riemann approximation step $m$ on the difference of trainable node positions determined by the Layer-Balanced Strategy ($q{=}10$). In particular, we compare the outcomes using $m{=}7$ and $m{=}20$ on Qwen2.5-0.5B-Instruct. Table~\ref{tab:qwen_sim_step} reports the similarity between the resulting trainable node positions under these two configurations.

To quantify the consistency of node selections between the two settings, we define similarity at both the layer level and the total network level. Let $A$ and $B$ denote the different sets of trainable node positions. Furthermore, let the network consist of $n$ layers, and $A_i \subset A$, $B_i \subset B$ denote the selected trainable node positions in the $i$-th layer under each setting.

The \textbf{layer-wise similarity} for the $i$-th layer is defined as:
\begin{equation}
\text{Sim}(A_i, B_i) = \frac{|A_i \cap B_i|}{|A_i|},
\end{equation}
which measures the proportion of nodes in $A_i$ that also appear in $B_i$.

The \textbf{overall similarity} across all layers is computed as the average of layer-wise similarities:
\begin{equation}
\text{Sim}(A, B) = \frac{1}{n} \sum_{i=1}^{n} \frac{|A_i \cap B_i|}{|A_i|}.
\end{equation}

\subsubsection{Number of samples}
\begin{table}[h!]
\centering
\begin{tabular}{c|c||c|c}
\toprule
\textbf{Layer} & \textbf{Sim. (\%)} & \textbf{Layer} & \textbf{Sim. (\%)} \\
\midrule
0  & 94.38  & 12 & 95.51 \\
1  & 94.38  & 13 & 96.63 \\
2  & 96.63  & 14 & 94.38 \\
3  & 94.38  & 15 & 92.13 \\
4  & 95.51  & 16 & 93.26 \\
5  & 92.13  & 17 & 94.38 \\
6  & 96.63  & 18 & 97.75 \\
7  & 96.63  & 19 & 94.38 \\
8  & 96.63  & 20 & 97.75 \\
9  & 96.63  & 21 & 96.63 \\
10 & 94.38  & 22 & 96.63 \\
11 & 92.13  & 23 & 93.26 \\
\midrule
\multicolumn{3}{r|}{\textbf{Total}} & \textbf{95.13} \\
\bottomrule
\end{tabular}
\caption{The similarity of each layer and the total similarity of trainable nodes obtained using two different MMLU sample selection methods on Qwen2.5-0.5B-Instruct.}
\label{tab:qwen_sim_sample}
\end{table}

We also investigated the impact of using different numbers of MMLU samples in the KVA process on the final trainable nodes. Specifically, we compared the similarity between using only the first 50 samples for each MMLU subset and using all MMLU samples. Note that we compare at $m{=}7$ and $q{=}10$. Following the definition of similarity in the previous section, we report the node similarity obtained by two methods on Qwen2.5-0.5B-Instruct and Llama3.1-8B-Instruct in Tables \ref{tab:qwen_sim_sample} and \ref{tab:llama3_sim_sample}, respectively.

\begin{table}[h!]
\centering
\begin{tabular}{c|c||c|c}
\toprule
\textbf{Layer} & \textbf{Sim. (\%)} & \textbf{Layer} & \textbf{Sim. (\%)} \\
\midrule
0  & 97.07  & 16 & 97.80 \\
1  & 97.07  & 17 & 98.53 \\
2  & 96.82  & 18 & 97.07 \\
3  & 97.31  & 19 & 96.58 \\
4  & 98.29  & 20 & 97.31 \\
5  & 97.07  & 21 & 98.29 \\
6  & 97.07  & 22 & 97.07 \\
7  & 97.56  & 23 & 97.56 \\
8  & 97.07  & 24 & 98.29 \\
9  & 97.80  & 25 & 98.04 \\
10 & 98.04  & 26 & 97.31 \\
11 & 97.56  & 27 & 98.78 \\
12 & 97.80  & 28 & 97.31 \\
13 & 97.56  & 29 & 98.04 \\
14 & 97.80  & 30 & 98.04 \\
15 & 97.56  & 31 & 96.82 \\
\midrule
\multicolumn{3}{r|}{\textbf{Total}} & \textbf{97.57} \\
\bottomrule
\end{tabular}
\caption{The similarity of each layer and the total similarity of trainable nodes obtained using two different MMLU sample selection methods on Llama3.1-8B-Instruct.}
\label{tab:llama3_sim_sample}
\end{table}

% \subsection{Prompts Used on MMLU}
% \subsection{Future Works}
\subsection{Discussions}
As mentioned in Section \ref{sec:Analysing}, the KVA process involves a one-time computational overhead, which can be considered from both the perspectives of time and computational cost. Here, we discuss potential directions for optimizing both types of overhead.

For the following two reasons, we believe that the current computational timing overhead can be significantly reduced:
(a) During the KVA process of Llama3.1-8B-Instruct on a single A100 GPU, we observed that the GPU was not fully utilized, with utilization ranging from approximately 50\% to 70\%. We attribute this underutilization to the fixed inference batch size being set equal to step $m$.
(b) The current implementation only supports single-GPU execution. We anticipate that extending the code to support multi-GPU execution will lead to a substantial reduction in computation time.

For the computational cost, as shown by our exploration of different values of m and sample sizes in this section, we believe there is still room for optimizing both the number of samples used for executing KVA on MMLU and the number of steps $m$ employed in the Riemann approximation.

In addition, we believe that exploring more suitable datasets for use in the KVA process is an appealing direction for future work.

\section{KVA Heatmaps}
% \subsection{Qwen2.5-0.5B-Instruct}

\begin{figure*}[h!]
    \centering
    \begin{subfigure}[c]{0.9\textwidth}
        \includegraphics[width=\linewidth]{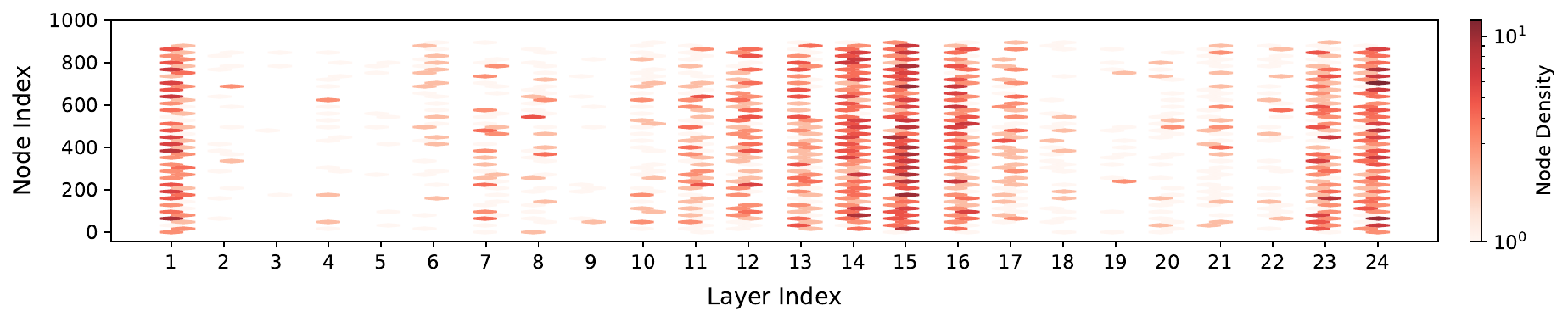}

        \label{fig:qwen_neuron_heatmap}
    \end{subfigure}
    \begin{subfigure}[c]{0.9\textwidth}
        \includegraphics[width=\linewidth]{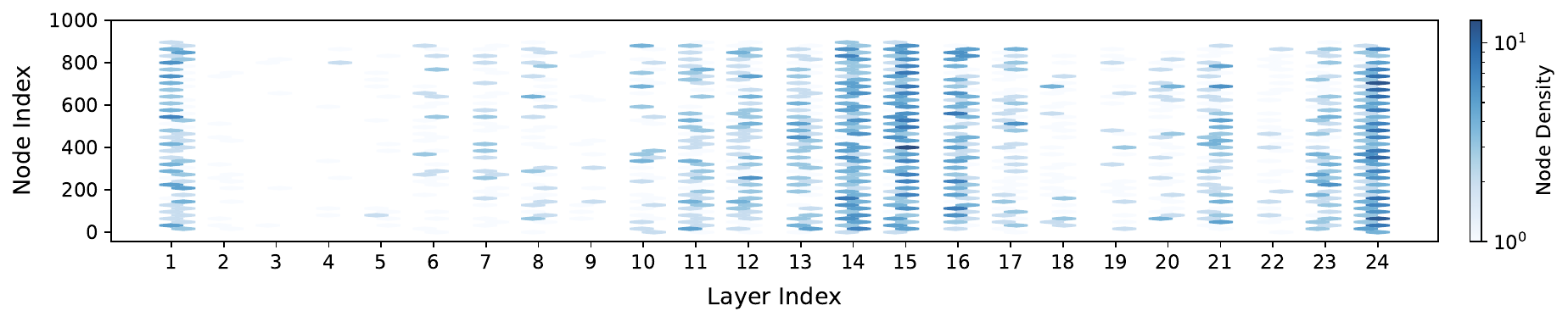}

        \label{fig:qwen_neuron_distribution}
    \end{subfigure}
    
    \caption{Heatmaps of top \(10\%\) KVA results across all 24 layers of Qwen2.5-0.5B-Instruct. The upper (red-tinted) map highlights the distribution of high-contribution node positions, while the lower (blue-tinted) map highlights the distribution of low-contribution node positions. Color intensity (log-scale) indicates the density of neurons in each category, with darker colors representing higher density.}

    \label{fig:kva_qwen}
    
\end{figure*}
% \subsection{Llama2-7B}

\begin{figure*}
    \centering
    \begin{subfigure}[c]{0.9\textwidth}
        \includegraphics[width=\linewidth]{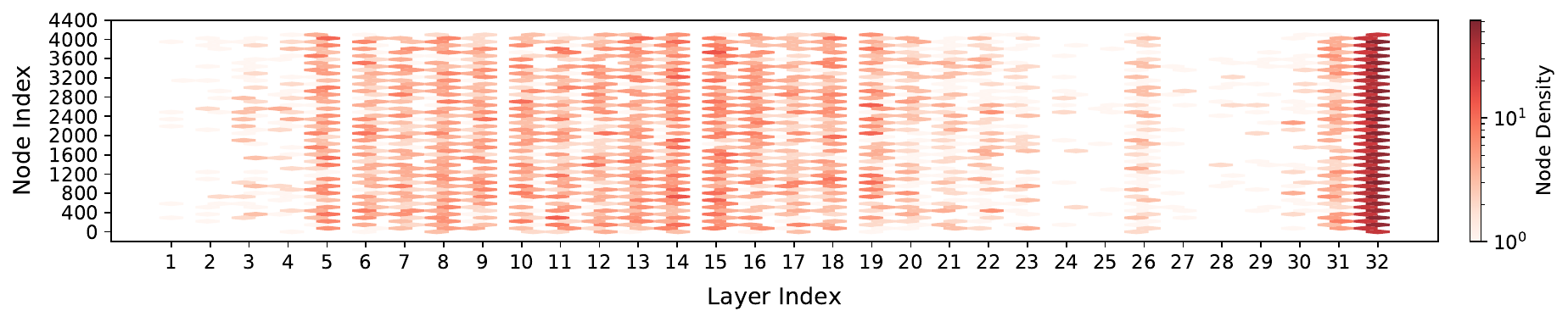}
        \label{fig:llama2_neuron_heatmap}
    \end{subfigure}
    
    \vspace{0.5em}
    
    \begin{subfigure}[c]{0.9\textwidth}
        \includegraphics[width=\linewidth]{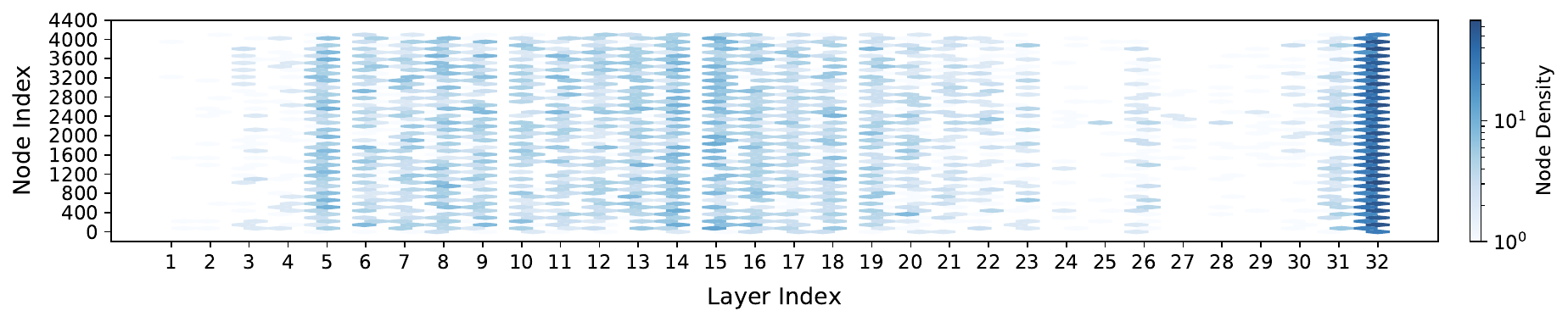}
        \label{fig:llama2_neuron_distribution}
    \end{subfigure}
    
    \caption{Heatmaps of top \(5\%\) KVA results across all 32 layers of Llama2-7B, following the same visualization settings as in Fig.~\ref{fig:kva_qwen}.}
    \label{fig:kva_llama2}
\end{figure*}

% \subsection{Llama3.1-8B-Instruct}
\begin{figure*}
    \centering
    \begin{subfigure}[c]{0.9\textwidth}
        \includegraphics[width=\linewidth]{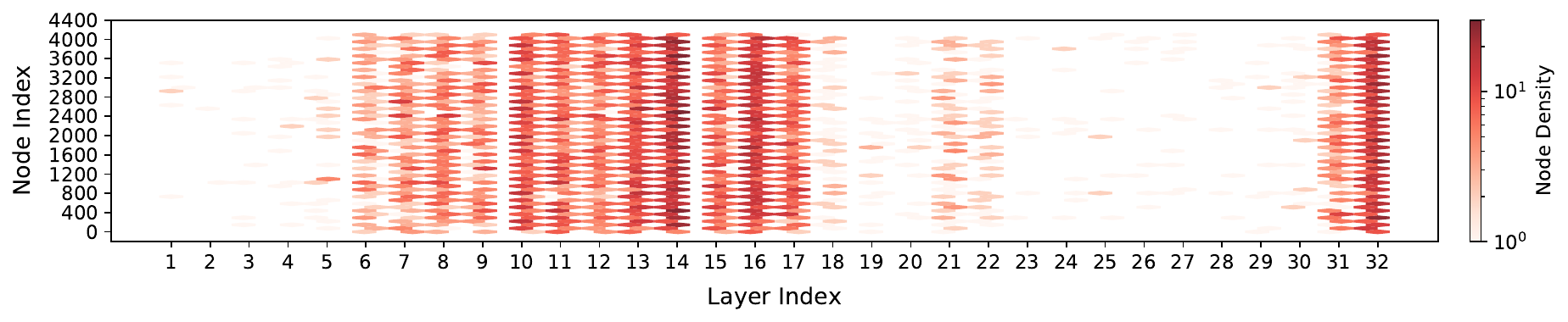}

        \label{fig:llama3_neuron_heatmap}
    \end{subfigure}
    \begin{subfigure}[c]{0.9\textwidth}
        \includegraphics[width=\linewidth]{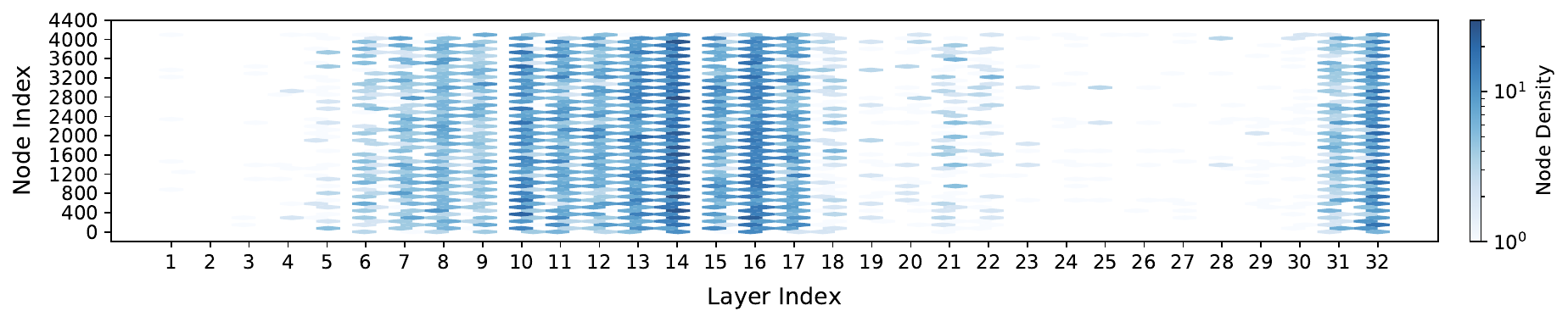}

        \label{fig:llama3_neuron_distribution}
    \end{subfigure}
    
    \caption{Heatmaps of top \(5\%\) KVA results across all 32 layers of Llama3.1-8B-Instruct, following the same visualization settings as in Fig.~\ref{fig:kva_qwen}.}

    \label{fig:kva_llama3}
    
\end{figure*}

\clearpage
\section{Implementation of LoKI Linear}
\label{sec:appendix_c}
We provide the PyTorch implementation code for LoKI Linear, and readers can access the complete project code on our public GitHub repository. Based on this implementation, we performed all the LoKI-based model training in the paper. Our code has been validated for use with Llama-Factory and supports integration with LoRA.

\begin{lstlisting}[language=Python, caption={PyTorch implementation of LoKI Linear}, label={lst:loki_linear}]
import torch
import torch.nn as nn


class LoKILinear(nn.Module):
    def __init__(self, original_linear, target_pos):
        super().__init__()
        self.out_features = original_linear.out_features
        self.in_features = original_linear.in_features
        self.active_pos = sorted(target_pos)
        self.frozen_pos = [
            i for i in range(self.out_features) if i not in self.active_pos
        ]

        # Parameter validation
        if not all(0 <= idx < self.out_features for idx in self.active_pos):
            raise ValueError(f"Activation indices must be within [0, {self.out_features - 1}]")
        if len(self.active_pos) != len(set(self.active_pos)):
            raise ValueError("Activation indices contain duplicate values")
        self.active = nn.Linear(self.in_features, len(self.active_pos), bias=False)
        self.frozen = nn.Linear(self.in_features, len(self.frozen_pos), bias=False)
        # Split the weight matrix
        W = original_linear.weight.data
        self.active.weight = nn.Parameter(W[self.active_pos].clone(), requires_grad=True)
        self.frozen.weight = nn.Parameter(W[self.frozen_pos].clone(), requires_grad=False)

        # Handle bias
        if original_linear.bias is not None:
            b = original_linear.bias.data
            self.active_bias = nn.Parameter(
                b[self.active_pos].clone(), requires_grad=True
            )
            self.frozen_bias = nn.Parameter(
                b[self.frozen_pos].clone(), requires_grad=False
            )
        else:
            self.register_parameter("active_bias", None)
            self.register_parameter("frozen_bias", None)

        # Pre-generate index mapping
        index_map = torch.empty(self.out_features, dtype=torch.long)
        index_map[self.active_pos] = torch.arange(len(self.active_pos))
        index_map[self.frozen_pos] = torch.arange(len(self.frozen_pos)) + len(
            self.active_pos
        )
        self.register_buffer("index_map", index_map)

    def forward(self, x):
        active_out = self.active(x)  # Compute active part via submodule
        frozen_out = self.frozen(x)    # Fixed part
        output = torch.cat([active_out, frozen_out], dim=-1)

        # Add combined bias
        if self.active_bias is not None:
            bias = torch.cat([self.active_bias, self.frozen_bias], dim=0)
            output += bias.unsqueeze(0).unsqueeze(0)  # Broadcast bias to all batches and sequence positions

        # Reorder output using pre-generated indices
        return output.gather(
            dim=-1,
            index=self.index_map.view(1, 1, -1).expand(
                output.size(0), output.size(1), -1
            ),
        )
\end{lstlisting}

\section{Experiment Setups for the LB Reranker Dataset}

\begin{table}[h!]
\centering
\small
\setlength{\tabcolsep}{2pt}
\resizebox{\columnwidth}{!}{%

\begin{tabularx}{\columnwidth}{l|c|c|c|c|c}
\toprule
Model & Learning Rate & \makecell{Batch\\Size} & Epochs & \makecell{LR\\Scheduler} & \makecell{WarmUp\\Ratio} \\ 
\midrule
LoKI($q{=}5$) & 1.0$\times$10$^{-5}$ & \multirow{4}{*}{1} & \multirow{4}{*}{1} & \multirow{4}{*}{cosine} & \multirow{4}{*}{0.01} \\ 
LoKI($q{=}10$) & 1.0$\times$10$^{-5}$ & & & & \\ 
LoKI($q{=}20$) & 5.0$\times$10$^{-6}$ & & & & \\ 
LoKI($q{=}30$) & 4.0$\times$10$^{-6}$ & & & & \\ 
\bottomrule
\end{tabularx}
}
\caption{Training hyperparameters on LB Reranker Dataset. LR denotes Learning Rate.}
\label{tab:rerank_param}
\end{table}

\begin{table}[h!]
\centering
\small
\begin{tabular}{l|c|c|c}
\toprule
Model   & Learning Rate & LoRA rank & LoRA alpha \\ 
\midrule
DoRA   & \multirow{3}{*}{1.0$\times$10$^{-5}$} & \multirow{3}{*}{8} & \multirow{3}{*}{16} \\
PiSSA  &                                       &                    &                     \\
CorDA  &                                       &                    &                     \\
\bottomrule
\end{tabular}
\caption{Training hyperparameters on LB Reranker Dataset when using DoRA, PiSSA, and CorDA. The parameters not mentioned in the table use the same parameters as in Table \ref{tab:rerank_param}.}
\label{tab:lb_sota}
\end{table}
On this dataset, we adopt the training parameter settings of the official full-parameter fine-tuning model, with the only modification being the learning rate. We find that when increasing the quota, appropriately lowering the learning rate helps maintain stable performance on general tasks while ensuring adaptability to downstream tasks. Table \ref{tab:rerank_param} shows the specific training hyperparameters for each model. Refer to the supplemental material for more detailed training hyperparameters. Due to variations in the availability of computing resources, we employed multiple hardware configurations to complete the training of these models. The configurations included the following setups: 8 RTX 4090 GPUs, 4 RTX 4090 GPUs, and 2 A100 GPUs. 

We use the BEIR evaluation code provided by the dataset provider to assess our models. Specifically, we evaluate on 9 datasets from BEIR: \texttt{Arguana}, \texttt{Dbpedia-entity}, \texttt{FiQA}, \texttt{NFCorpus}, \texttt{SCIDOCS}, \texttt{SciFact}, \texttt{TREC-COVID-v2}, \texttt{ViHealthQA}, and \texttt{Webis-Touche2020}. For each dataset, we evaluate on a subset of the queries (the first 250).
\section{Experiment Setups for the ToolACE Function-Calling Dataset}
\begin{table}[h!]
\centering
\small
\setlength{\tabcolsep}{2pt}
\begin{tabularx}{\columnwidth}{l|c|c|c|c|c}
\toprule
Model & Learning Rate & \makecell{Batch\\Size} & Epochs & \makecell{LR\\Scheduler} & \makecell{WarmUp\\Ratio} \\ 
\midrule
LoKI($q{=}10$) & \multirow{3}{*}{9.0$\times$10$^{-6}$} & \multirow{3}{*}{4} & \multirow{3}{*}{3} & \multirow{3}{*}{cosine} & \multirow{3}{*}{0.1} \\ 
LoKI($q{=}20$) & & & & & \\ 
LoKI($q{=}30$) & & & & & \\ 
\bottomrule
\end{tabularx}
\caption{Training hyperparameters on ToolACE Function-Calling Dataset. LR denotes Learning Rate.}
\label{tab:toolace_s_params}
\end{table}

\begin{table}[h!]
\small
\centering
\begin{tabular}{l|c|c|c}
\toprule
Model           & Learning Rate  & LoRA rank   & LoRA alpha \\ 
\midrule
LoKI*($q{=}30$) & 5.0$\times$10$^{-4}$ &        32   &      64      \\ 
\bottomrule
\end{tabular}
\caption{Training hyperparameters on ToolACE Function-Calling Dataset for LoKI*($q{=}30$). The parameters not mentioned in the table use the same parameters as in Table \ref{tab:toolace_s_params}.}
\label{tab:toolace_l_params}

\end{table}

\begin{table}[h!]
\centering
\begin{tabular}{l|c|c|c}
\toprule
Model   & Learning Rate     & LoRA rank & LoRA alpha \\ 
\midrule
DoRA   & \multirow{2}{*}{1.0$\times$10$^{-4}$} & \multirow{2}{*}{16} & \multirow{2}{*}{32} \\
PiSSA  &                                &                &               \\
\bottomrule
\end{tabular}
\caption{Training hyperparameters on ToolACE Function-Calling Dataset when using DoRA and PiSSA. The parameters not mentioned in the table use the same parameters as in Table \ref{tab:toolace_s_params}.}
\label{tab:toolace_sota_param}
\end{table}

Apart from the learning rate and batch size, we adopt the training hyperparameters reported by the official model provider. Table \ref{tab:toolace_s_params} presents the training parameters used in our standard training approach, while Table \ref{tab:toolace_l_params} reports the hyperparameters used in the LoRA-integrated training setup. Refer to the supplemental material for complete training configuration files. All models trained on this dataset were completed using 2 A100 GPUs.

As mentioned in the main text, we use the publicly available evaluation codebase for the Berkeley Function Calling Leaderboard (BFCL).

\section{Benchmark Execution Details}
To ensure a fair comparison of model performance, we use OpenCompass to evaluate all the models discussed in this paper, including the two base models: Llama3.1-8B-Instruct and Qwen2.5-0.5B-Instruction.
The version of the benchmarks can be viewed in Table \ref{tab:dataset_metrics}. Specifically, for IFEval, we calculate the average of four indicators. The scores for the remaining benchmarks are reported directly.

\begin{table}[h]
\centering
\small
\setlength{\tabcolsep}{2pt}

\begin{tabularx}{\columnwidth}{l|r|r|r}
\toprule
\textbf{Dataset} & \textbf{Version} & \textbf{Metric} & \textbf{Mode} \\
\midrule
winogrande & 458220 & accuracy & gen \\
openai\_humaneval & 8e312c & humaneval\_pass@1 & gen \\
gsm8k & 1d7fe4 & accuracy & gen \\
triviaqa & 2121ce & score & gen \\
hellaswag & 6faab5 & accuracy & gen \\
IFEval & 3321a3 & Prompt-level-strict-accuracy & gen \\
IFEval & 3321a3 & Inst-level-strict-accuracy & gen \\
IFEval & 3321a3 & Prompt-level-loose-accuracy & gen \\
IFEval & 3321a3 & Inst-level-loose-accuracy & gen \\
\bottomrule
\end{tabularx}
\caption{Specific information on all benchmarks in OpenCompass.}
\label{tab:dataset_metrics}
\end{table}

\section{Details for Ablation Studies}

\begin{table}[h]

\setlength{\tabcolsep}{1pt}  % 可选:控制列间距
\centering
{\small
\begin{tabularx}{\columnwidth}{l|*{6}{>{\centering\arraybackslash}X}|r}
\toprule
Model & \makecell{Trivia\\QA} & \makecell{GSM\\8K} & \makecell{Hella\\Swag} & \makecell{Wino\\Grande} & \makecell{Human\\Eval} & \makecell{IF\\Eval} & \makecell{Avg(↓)} \\
\midrule
LoKI    & 20.53 & 37.98 & 21.53 & 47.04 & 24.39 & 33.39 & 8.86\%\\
\midrule
\makecell{G-H\\($q{=}10$)} & 11.04 & 29.34 & 10.23 & 17.68 & 25.00 &  26.79 & 39.04\%\\
\makecell{G-L\\($q{=}10$)} & 9.83 & 30.93 & 8.48 & 43.09 & 24.39 &  27.72 & 30.48\%\\
\bottomrule
\end{tabularx}
}
\caption{Performance comparison of different methods on general task benchmarks.}\label{tab:ablation_lbs_full}
\end{table}

\begin{table}[h]

\setlength{\tabcolsep}{1pt}

\centering
{\small
\begin{tabularx}{\columnwidth}{l|*{6}{>{\centering\arraybackslash}X}|r}
\toprule
Model & \makecell{Trivia\\QA} & \makecell{GSM\\8K} & \makecell{Hella\\Swag} & \makecell{Wino\\Grande} & \makecell{Human\\Eval} & \makecell{IF\\Eval} & \makecell{Avg(↓)} \\
\midrule
Llama     & 65.77  & 84.46  & 73.85 & 62.98 & 68.29 & 79.76 & NaN \\
\midrule
S-H & 62.32 & 15.16 & 23.54 & 60.93 & 43.29 & 59.73 & 36.73\%  \\
S-L & 62.05 & 57.85 & 73.84  & 60.3 & 54.88 & 74.15 & 11.35\%  \\
\bottomrule 
\end{tabularx}
}
\caption{Performance of two suppression strategies on benchmarks when $q{=}1$.}\label{tab:ablation_kva_full}
\end{table}

\begin{table}[h]

\centering
{\scriptsize
\setlength{\tabcolsep}{2pt}
\begin{tabularx}{\columnwidth}{l|*{8}{>{\centering\arraybackslash}X}}
\toprule
Model & \makecell[c]{MAP\\@1\\(\%)} & \makecell[c]{MAP\\@10\\(\%)} & \makecell[c]{Recall\\@1\\(\%)} & \makecell[c]{Recall\\@10\\(\%)} & \makecell[c]{NDCG\\@1\\(\%)} & \makecell[c]{NDCG\\@10\\(\%)} & \makecell[c]{P\\@1\\(\%)} & \makecell[c]{P\\@10\\(\%)} \\
\midrule
\makecell{G-H\\($q{=}10$)} & -1.0 & -2.8 & -1.0 & -1.6 & 0.4 & -1.2 & 0.0 & -1.0 \\
\makecell{G-L\\($q{=}10$)} & -4.5 & -4.8 & -4.5 & -2.8 & -2.6 & -3.0 & -2.6 & -3.0 \\
\bottomrule
\end{tabularx}
}
\caption{Comparison with the full parameter fine-tuning model on BEIR. Table shows the percentage difference in performance compared to the full parameter fine-tuning model across standard retrieval metrics.}\label{tab:beir}
\end{table}

\section{Limitations and Future Works}
While our method demonstrates competitive performance across experimental evaluations, several limitations warrant discussion. First, the effectiveness of LoKI is influenced by the hyperparameter $q$, which controls the proportion of trainable nodes. While we provide empirical guidelines (e.g., $q{=}10\text{--}30$), optimal values vary across tasks and models. A more robust, adaptive mechanism for quota allocation would improve LoKI's general usability. Second, the one-time computational overhead introduced by KVA has significant potential for optimization, especially in implementing code. As shown in Appendix \ref{appendix_a}, there is still a trade-off to be addressed between specific parameter settings and overall performance. Additionally, when integrating LoRA within the LoKI framework, the interaction between LoRA training parameters and the LoKI $q$ value remains underexplored. Finally, the attribution techniques employed in the analysing phase offer further opportunities for enhancement. Future work could focus on improving both the accuracy and efficiency of these attribution methods.

\end{document}